\documentclass{article} 
\usepackage{iclr2026_conference,times}

\usepackage{hyperref}
\usepackage{url}
\usepackage{amsmath}
\usepackage{newtxtext,newtxmath}
\usepackage{booktabs}
\usepackage{float}      
\usepackage{array}
\usepackage{multirow}
\usepackage{enumitem}
\usepackage{graphicx}
\graphicspath{{figs/}{./}}

\title{SkillEvo: Self-Renewing Evolution Gradients \\ from Multi-Turn Interaction Feedback}

\author{%
  Qianxi Yan$^{\dagger,\ddag}$\thanks{Corresponding author.} \quad
  Chunrong Chen$^{\dagger}$ \quad
  Jiuzhou Zhao$^{\dagger}$ \quad
  Min Zhang$^{\dagger}$ \\[0.4em]
  \textbf{Yongzhou Xu}$^{\dagger}$ \quad
  \textbf{Xiaochuan Xu}$^{\dagger}$ \\[0.4em]
  \begin{tabular}[t]{>{\centering\arraybackslash}p{0.85\textwidth}}
   $^{\dagger}$Tencent Cloud Andon \quad $^{\ddag}$Zhejiang University \\[0.3em]
   \small
   Qxxx2616@zju.edu.cn \\[0.15em]
   \{charentchen, joskazhao, alexzmzhang, alanxu, xxcxu\}@tencent.com
  \end{tabular}%
}
\iclrfinalcopy 
\begin{document}

\maketitle
\begin{abstract}
Agent Skills are today either hand-authored or produced in a single LLM generation pass, and consequently possess no closed loop through which they might improve from the interaction failures they actually cause. Recent work does close this loop, but derives its feedback from single-turn question-answering evaluation. The consequence is a sharp asymmetry: once the first round has patched the gaps that a single exchange can reveal, the evolution gradient decays, the defects that surface only across multiple turns remain invisible, and evolution stalls. Governance in these systems is likewise driven by an end-to-end verification score---a scalar gate that can reject a degraded candidate but can neither localize nor repair its structural cause. We argue that the binding constraint on sustained skill evolution is neither editing capability nor the number of iterations, but whether the evaluation feedback keeps supplying \emph{trustworthy evolution gradients}. We introduce \textbf{SkillEvo}, in which trustworthy feedback generates the gradient and controllable governance constrains its direction. The first component recasts multi-turn user simulation from an evaluation endpoint into a feedback generator: follow-up questions expose defects layer by layer, so that every round of revision both consumes feedback and produces new feedback. The second replaces the passive rejection of a scalar gate with an independent governance layer that actively repairs factual degradation and structural bloat, preventing the gradient from drifting as degradation accumulates. Across six categories of cloud services, 9 production Skills, and 98 skill-reference files, SkillEvo surpasses self-reflection-based evolution by 23.0 points and single-turn-QA-driven evolution by 15.4 points.
\end{abstract}

\fancyhead{}\fancyfoot{}
\setlength{\headheight}{24pt}
\renewcommand{\headrulewidth}{0.4pt}
\fancyhead[L]{\raisebox{-0.5\height}{\includegraphics[height=14pt]{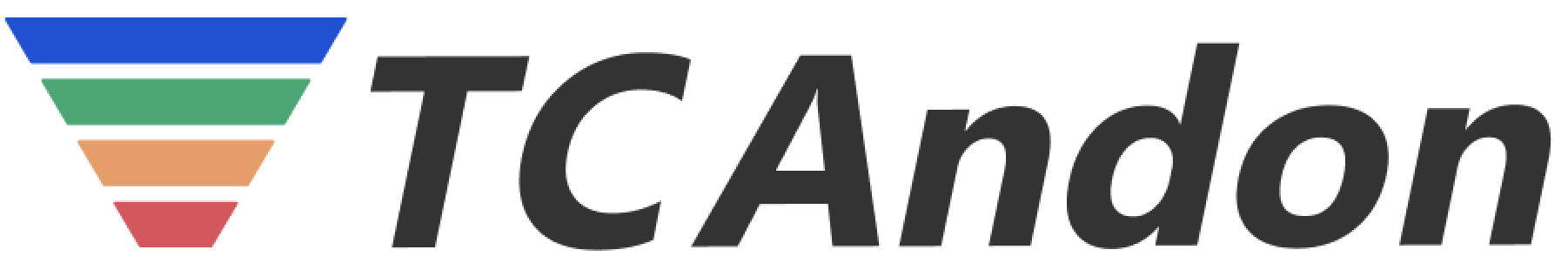}}}

\begin{figure}[t]
\centering
\makebox[\textwidth][c]{\includegraphics[width=1.08\textwidth]{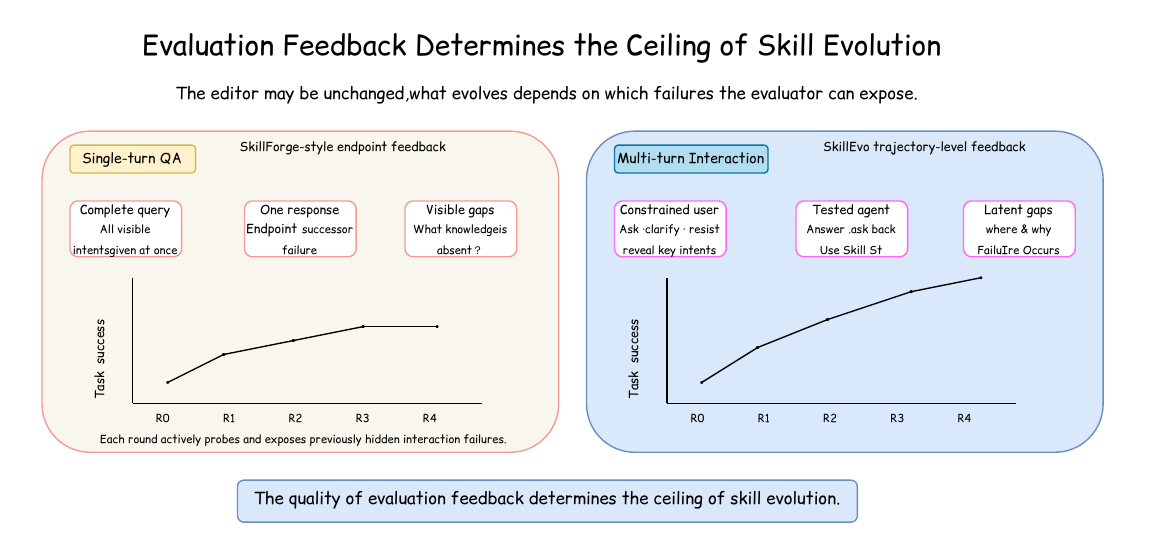}}
\caption{The evaluation feedback modality determines the ceiling of skill evolution. \textbf{Left:} single-turn QA patches the gaps visible in the first exchange; the evolution gradient then decays and the TSR curve saturates around round 2. \textbf{Right:} multi-turn interaction keeps exposing latent failures---each round of revision lets the dialogue proceed further and reach the next layer of defects---so the gradient renews itself and TSR keeps climbing.}
\label{fig:trajectory}
\end{figure}

\section{Introduction}

Language agents are moving beyond single-turn question answering toward autonomy: multi-turn interaction, knowledge retrieval, and tool invocation~\citep{wang2024survey}. In the cloud-services domain, RCACopilot~\citep{chen2024automatic} applies LLMs to incident root-cause analysis, RCAgent~\citep{wang2024rcagent} endows the agent with autonomous diagnosis, and D-Bot~\citep{zhou2023d} structures database diagnosis via tree search; all three, however, focus on single-exchange capability and do not address the long-term bottleneck of customer support: the continual maintenance of Skills---portable modules that encapsulate domain knowledge and handling procedures. Skill maintenance still relies on manual authoring, and failed interactions cannot be consolidated automatically into reusable knowledge. Existing continual-improvement research~\citep{zhao2025agent} depends on human-annotated feedback, which is costly and narrow in coverage. Converting failed interactions in real tickets into feedback that drives continual Skill improvement---automatically, and in a closed loop---is therefore the key to escaping the scale and latency limits of manual maintenance.

Our central claim is that \textbf{the quality of Skill self-evolution is governed by the quality of the feedback signal and by controllable evolution governance, rather than by editing capability or the number of iterations}. The first requirement is feedback that both reveals the latent defects arising in multi-turn interaction and can be attributed to repairable knowledge gaps. The second is a revision sequence that converges: fact consistency and structural consistency must be maintained so that degradation does not accumulate. Existing work is structurally deficient on both counts---its feedback is confined to what a single exchange makes visible, and its governance rests on a scalar gate that can reject a degraded result but cannot identify its structural cause. Figure~1 contrasts the two feedback modalities: the single-turn gradient saturates once the first round has patched the visible gaps, whereas multi-turn interaction keeps renewing it.

We therefore propose \textbf{SkillEvo}, a skill-evolution framework resting on two pillars. Trustworthy feedback recasts multi-turn user simulation from an evaluation endpoint into a feedback generator, so that follow-up questions keep generating fresh gradients. Controllable governance treats a Skill as a structured knowledge system and has an independent layer actively repair factual degradation and structural bloat after each revision, keeping the gradient direction from drifting as degradation accumulates.

Our contributions are as follows.

\begin{enumerate}[leftmargin=*,itemsep=3pt,parsep=0pt,topsep=3pt]
\item We formalize three necessary conditions for trustworthy feedback---coverage, accuracy, and attributability---and recast multi-turn user simulation from an evaluation endpoint into a feedback generator. An intent state machine gates coverage; dual-sided orthogonal evaluation partitions responsibility between simulator and service agent to secure accuracy; and collective attribution screens repairable gaps by root cause while distilling their cross-sample commonalities. Together they furnish a trustworthy gradient for skill evolution.
\item We identify a Skill as a structured knowledge system whose multi-round revision incurs degradations that a scalar score cannot diagnose: knowledge bloat, reference breakage, and factual over-generalization. SkillEvo maintains fact consistency against dual anchors and repairs all three through graph-structural diagnosis, shifting governance from scalar-gated passive rejection to diagnosis-driven active repair.
\item Across six categories of cloud services, 9 Skills, and 98 skill-reference files, SkillEvo improves over the original Skills by 51.8 points, over self-reflection-based evolution by 23.0 points, and over single-turn-QA-driven evolution by 15.4 points. The framework is deployed in the Tencent Cloud production environment, evidencing its effectiveness under real operating conditions.
\end{enumerate}

\section{Related work}

\subsection{User simulation}

User simulation is now standard for the automatic evaluation of task-oriented dialogue. $\tau$-bench~\citep{yao2024tau} advances evaluation from ``can the task be completed'' to ``can it be completed \emph{reliably}'' by grounding tool-agent-user interaction in real-world domains. ECom-Bench~\citep{wang2025ecom} and VoiceAgentEval~\citep{xu2025voiceagenteval} raise fidelity by grounding the simulator in real user personas, while SAGE~\citep{shea2026sage} injects business profiles and enterprise knowledge bases into the simulator and generates probing questions in reverse, surfacing 33\% more agent errors than a generic user.

These results establish that simulation exposes errors invisible to single-turn testing, yet they share one limitation: \textbf{all of them treat simulation as the evaluation endpoint}. A trajectory is discarded once it has served to adjudicate success or failure; failure cases are neither separated into repairable gaps and non-Skill limitations, nor returned to the Skill---evaluation and evolution remain disjoint. The reliability of the simulation itself is moreover left unexamined: whether the simulator adequately raises the user's real intents is never verified, and simulation distortion contaminates the evaluation conclusion directly. SkillEvo departs from this practice by recasting multi-turn simulation into a feedback generator for evolution: an intent state machine gates coverage, dual-sided orthogonal evaluation isolates distortion, and independent attribution screens repairable gaps and returns them to the Skill. Concurrently, SEAD~\citep{dai2026sead} also grounds service-agent evolution in multi-turn dialogue, but optimizes model parameters via reinforcement learning rather than evolving a textual Skill knowledge base.

\subsection{Agent skill evolution}
\label{sec:related_evo}

Agent Skills~\citep{zhang2025agentskills} have emerged as a decisive carrier of agent capability, empirically validated as a unified evaluation surface~\citep{li2026skillsbench} and surveyed as a knowledge-system category beyond tool use~\citep{jiang2026sok}, yet their maintenance still rests on manual authoring. Existing self-evolution work~\citep{gao2025survey} falls into three tiers by feedback source, all of them structurally deficient at the levels of feedback signal and evolution governance.

\textbf{Feedback signal.} Self-Refine~\citep{madaan2023self} reflects and edits without any evaluation; unable to separate genuine gaps from the model's own blind spots, it does not constitute an evolution sequence. SkillForge~\citep{liu2026skillforge} and the broader family of single-turn QA evaluation methods~\citep{yang2026skillopt,alzubi2026evoskill,ni2026trace2skill,chen2026skillcat,agrawal2026gepa,yuksekgonul2024textgrad} capture only single-turn-visible gaps, so their gradient decays after the first round; worse, their feedback drives revision without attribution screening, so irreparable signals are mis-encoded as knowledge and produce document bloat and factual conflict. Multi-turn evaluation such as $\tau$-bench~\citep{yao2024tau} and SAGE~\citep{shea2026sage} does expose interaction-level defects, but stops at the evaluation endpoint; no prior work returns multi-turn trajectories to the Skill. Concurrent skill-evolution efforts---SkillFoundry~\citep{shen2026skillfoundry}, SkillX~\citep{wang2026skillx}, EvoSkills~\citep{zhang2026coevoskills}, AgentSkillOS~\citep{li2026organizing}, and Steve-Evolving~\citep{xie2026steve}---likewise acquire skills from execution traces or co-evolution signals, not from the layered defects that follow-up questions expose.

\textbf{Evolution governance.} Even trustworthy feedback does not preclude degradation under successive revision, and existing governance~\citep{yang2026skillopt,chen2026skillcat} operates on a single text file paired with a scalar score. Even work that explicitly raises governance concerns---co-evolutionary verification~\citep{zhang2026coevoskills}, RL-based skill curation~\citep{ouyang2026skillos}, and audited skill-graph self-improvement~\citep{huang2025audited}---still gates candidates by pass/fail or scalar reward, without diagnosing the structural degradation of a textual knowledge base. Degradation is thus inferred only indirectly from a falling total, the score cannot localize its cause, and the gate can reject an entire candidate but never repair the offending structure. The append-only policy of SkillForge~\citep{liu2026skillforge} compounds the problem by letting the Skill grow without bound. Once a Skill is a multi-file directed graph of routing table and references, degradations such as dangling references, orphan files, and factual over-generalization become \textbf{inexpressible in scalar score space}, and can therefore be neither diagnosed nor repaired in a targeted manner.

SkillEvo closes both gaps. At the feedback level it is, to our knowledge, the first framework to return multi-turn simulated dialogue to skill evolution: an intent state machine guarantees coverage, dual-sided orthogonal evaluation isolates simulation distortion, and collective attribution screens repairable gaps, recasting multi-turn interaction from an evaluation endpoint into a feedback generator. At the governance level, dual anchors maintain fact consistency as a hard constraint while graph-structural diagnosis repairs knowledge bloat, reference breakage, and factual over-generalization as a soft constraint, moving governance from scalar-gated passive rejection to diagnosis-driven active repair.

\section{Method}

\begin{figure}[ht] \centering \includegraphics[width=\textwidth]{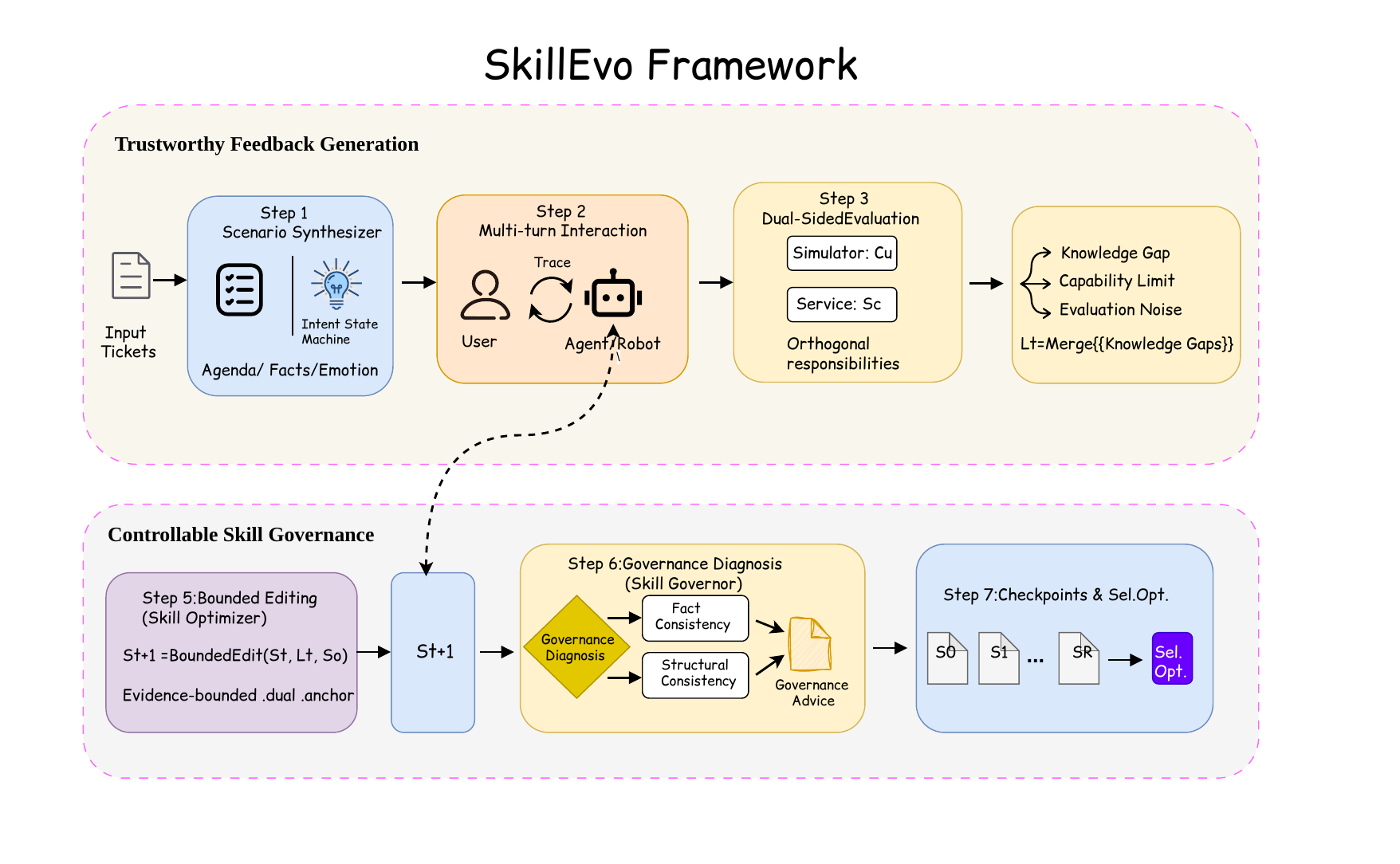} \caption{Overview of SkillEvo. The upper layer generates trustworthy feedback to drive revision; the lower layer governs structural degradation so that the knowledge carrier remains intact.} \label{fig:framework} \end{figure}

\subsection{Overall framework}

Let $U$ denote a task-constrained User Agent, $S_t$ the Skill in use at round $t$, and $\pi(S_t)$ the service agent loaded with that Skill. SkillEvo evolves the Skill through the closed loop $\text{Scenario Synthesizer} \to \text{UserAgent} \to \text{Verifier} \to \text{Collective Attribution} \to \text{Skill Optimizer} \to \text{Skill Governor}$ (Figure~2).

\begin{enumerate}[leftmargin=*,itemsep=1pt]
\item \textbf{Scenario Synthesizer.} A constrained user scenario is synthesized from a real human-handled ticket by extracting the intent agenda, behavior facts, emotion trajectory, and human reference solution, which together constitute the evaluation task.
\item \textbf{User Agent.} $U$ interacts with $\pi(S_t)$ over multiple turns, producing the trajectory $\tau_t = \operatorname{Interact}(U, \pi(S_t))$.
\item \textbf{Verifier.} The trajectory is judged against the human reference solution to determine success or failure and to generate feedback $(r_t, f_t) = \operatorname{Verify}(\tau_t)$, where $r_t \in \{\text{Success}, \text{Failure}\}$ and $f_t$ carries the failure cause and supporting evidence.
\item \textbf{Collective Attribution.} Failures are classified by repairability into $a_t \in \{\text{Knowledge Gap},\ \text{Capability Limit},\ \text{Evaluation Noise}\}$; only Knowledge Gap is projected into the evaluation feedback $\mathcal{L}_t$.
\item \textbf{Skill Optimizer.} $\mathcal{L}_t$ drives a bounded update of the Skill: $S_{t+1} = \operatorname{Update}(S_t, \mathcal{L}_t, S_0)$. ``Bounded'' has two layers of meaning: an \emph{evidence boundary}---only the verified gaps in $\mathcal{L}_t$ are patched, and no unsupported content is introduced---and a \emph{reference boundary}---revision is anchored to the production baseline $S_0$ so that new knowledge never overwrites existing stable facts.
\item \textbf{Skill Governor.} After revision, an independent governance layer detects structural degradation (knowledge bloat, reference breakage, factual over-generalization) and emits recommendations that are merged with the attribution signals, so that the next round repairs structure and supplements knowledge simultaneously.
\end{enumerate}

\subsection{Trustworthy feedback}

A feedback signal $\mathcal{L}_t$ is trustworthy if and only if three conditions hold simultaneously: \textbf{coverage} ($c_U=1$; all key intents have been raised and the simulation is faithful), \textbf{accuracy} ($s_C$ judges each exposed intent individually against the human reference), and \textbf{attributability} (the failure's root cause is repairable by the Skill). We address these in turn as intent coverage, responsibility separation, and repairability screening.

\subsubsection{Trustworthy user reconstruction}
\label{sec:synth}

From the complete dialogue trace of each human-handled ticket we extract three kinds of information to reconstruct a constrained user:

\begin{enumerate}[leftmargin=*,itemsep=2pt,parsep=0pt,topsep=2pt]
\item \textbf{Intent agenda}: the user's core requests and optional follow-up questions are extracted from the ticket and labeled \texttt{key}/\texttt{minor}, fixing what the simulated user ought to raise.
\item \textbf{Behavior facts}: the information the user already holds, the operations already attempted, and the observable symptoms are injected into the User Agent, constraining its replies to real information rather than fabrication.
\item \textbf{Emotion trajectory}: fluctuations in cooperativeness and emotional turning points are modeled so that the user changes pace realistically and can reasonably refuse when the agent requests complex operations.
\end{enumerate}

What this reconstruction targets is \emph{task-level constrained realism}: the agenda fixes the real requests, behavior facts bound the information surface, and the emotion trajectory constrains the interaction pace. The goal is to prevent intent omission, factual fabrication, and answer leakage---not to replicate the full linguistic distribution of a real user.

\textbf{Intent state machine.} The intent state machine tracks, for each intent, whether it \emph{has been raised} and whether it \emph{has been substantively addressed}. Normal termination is permitted only once all intents satisfy both conditions; when the agent cannot resolve the issue or repeatedly stalls, an abandonment-style termination is triggered instead. This rules out premature stopping and redundant turns alike.

\subsubsection{Dual-sided orthogonal evaluation}
\label{sec:dual}

SkillEvo evaluates the simulator side and the agent side separately, rendering responsibility separable: a failure occurring where an intent was never raised is attributed to simulation distortion rather than to the agent's Skill.

\textbf{Simulator side: intent coverage.} Let $\mathcal{K}$ be the set of key intents in the agenda and $\mathcal{K}_{\text{asked}}$ the subset actually raised; intent coverage is $c_U = |\mathcal{K}_{\text{asked}}|/|\mathcal{K}|$. Whenever $c_U < 1$, the sample is excluded from the agent-side denominator and attributed to Evaluation Noise.

\textbf{Agent side: Skill hit and exposed-intent response accuracy.} Skill hit acts as a gating factor: $h=1$ on a hit, and $h=0$ otherwise, which forces $s_C=0$ and an immediate failure. Each raised intent is judged against the human ``intent--solution'' pairs with $a_i \in \{0,1\}$ and weighted by priority ($\alpha=0.7$ for key, $1-\alpha$ for minor):
\begin{equation}
s_C = h \cdot \frac{\sum_{i} w_i\, a_i}{\sum_{i} w_i}, \quad w_i = \begin{cases} \alpha, & \text{priority}_i = \text{key} \\ 1-\alpha, & \text{priority}_i = \text{minor} \end{cases}
\end{equation}
We stress that $s_C$ is an intent-level quantity: it constrains the reliability of judgments on already-exposed intents, not the end-to-end resolution rate. A low $s_C$ is precisely what supplies the evaluation feedback.

\subsubsection{Collective attribution}
\label{sec:attr}

A low score does not by itself imply that the Skill lacks knowledge; the failure may equally stem from permission or tooling limits, or from an unreliable evaluation. The Attributor compares the human handling process, the simulated dialogue, and the Verifier's evidence~\citep{wei2022chain,yao2023tree,zhang2025agent}, and classifies failures by \textbf{whether the Skill can repair them}: \textbf{Knowledge Gap}, where a stable fact present in the human reference was omitted or answered incorrectly; \textbf{Capability Limit}, covering permission and tooling restrictions, poor delivery, and infrastructure faults; and \textbf{Evaluation Noise}, covering false negatives and scenario distortion. Only Knowledge Gap is projected into the feedback; the remaining two classes are isolated from the revision loop.

Because multiple failures within a round frequently point to the same gap, SkillEvo merges all Knowledge Gap cases by semantic similarity into a single feedback signal, $\mathcal{L}_t = \operatorname{Merge}(\{f \in \mathcal{F}_t \mid a_t(f) = \text{Knowledge Gap}\})$, focusing revision on cross-sample commonalities instead of per-instance noise.

\subsection{Controllable skill self-evolution}

Bounded revision delimits the content boundary of a single round, but multi-round revision accumulates degradation at the level of the sequence: new knowledge may be introduced at the cost of losing existing stable facts, and revision may fracture the Skill's structural integrity. SkillEvo therefore charges an independent governance layer with maintaining fact consistency, a hard constraint that rejects violating candidates, and structural consistency, a soft constraint that emits recommendations to drive subsequent repair.

\subsubsection{Fact consistency}

Fact consistency requires a revised Skill to retain at least the stable facts of the production baseline: $\operatorname{Facts}(S_t) \supseteq \operatorname{Facts}(S_0) \cap \mathcal{S}_{\text{stable}}$. The constraint is checked by an independent inspector against \emph{dual anchors}: $S_0$ detects fact loss accumulated across rounds, and $S_{t-1}$ detects factual errors newly introduced in the current round, so that degradation is attributed to the correct revision source. A single anchor at $S_0$ cannot distinguish the two, which leaves the repair direction ambiguous.

Three classes of violation are detected: \textbf{knowledge loss} against $S_0$, where a stable fact has been deleted; \textbf{process errors} against $S_{t-1}$, comprising factual errors newly introduced this round; and \textbf{self-contradiction} globally, where the revised Skill asserts conflicting statements. Any violation rejects the candidate and triggers same-round repair, which consults both diffs: deleted lines in $S_0 \to S_t$ localize the lost facts to restore, while changed lines in $S_{t-1} \to S_t$ forbid reverting to the original. Repair thus restores lost facts while preserving legitimate new knowledge.

\subsubsection{Structural consistency}
\label{sec:gov}

The knowledge organization of a Skill forms a directed graph in which routing nodes point to knowledge nodes and knowledge nodes cross-reference one another. Successive revision erodes this graph along three dimensions, none of which a scalar score can perceive:

\begin{itemize}[leftmargin=*,itemsep=1pt]
\item \textbf{Knowledge bloat}: redundancy grows within nodes, diluting routing precision.
\item \textbf{Reference breakage}: inter-node connectivity is severed, as dangling references or orphan files.
\item \textbf{Factual over-generalization}: concrete values, versions, and rules decay into vague statements, weakening the usability of answers.
\end{itemize}

Unlike fact consistency, which rejects candidates outright, structural consistency operates as a soft constraint: rather than discarding the revision, it merges governance recommendations with the attributed gaps and injects them into the next round, so that structural degradation is dissolved round by round along the iteration sequence.

\section{Experiments}

\subsection{Experimental setup}

\subsubsection{Scenarios and dataset}

We evaluate SkillEvo on the production technical-support scenarios of Tencent Cloud, spanning six categories of cloud services, 9 production Skills, and 98 skill-reference files (Appendix~A, Table~7). Every ticket in the dataset was escalated to a human agent: roughly 40\% at the very beginning of the interaction and roughly 60\% after several unresolved rounds. The dataset is thus the \emph{failure set} of the existing Skills---each ticket corresponds to a knowledge gap that a real user has already exposed and the current Skill fails to cover.

Within each scenario, the tickets of each Skill are ordered chronologically and split into four equal parts. The first three constitute the development set and drive the evolution loop: scenario synthesis, simulated interaction, attribution, and revision touch only this set. The fourth is held out as the evaluation set, is fed back into no stage of the loop, and serves solely for measurement and reporting. All TSR values reported below are measured on the evaluation set, whereas version selection relies exclusively on the development set.

\subsubsection{Baselines and variants}

\begin{table}[ht]
\centering
\small
\caption{Baselines and variants.}
\label{tab:baselines}
\begin{tabular}{p{2.9cm}p{6.2cm}cc}
\toprule
\textbf{Method} & \textbf{Skill source and update mechanism} & \textbf{Multi-turn} & \textbf{Rounds} \\
\midrule
Original Skill & Hand-authored initial Skill, never updated & No & 0 \\
Self-Reflection & Model self-reflects and edits the Skill directly, no evaluation feedback & No & 4 \\
Single-turn QA & Single-turn QA evaluation-driven evolution~\citep{liu2026skillforge} & No & 4 \\
\textbf{SkillEvo} & Iterative evolution driven by simulated multi-turn interaction & Yes & 4 \\
\bottomrule
\end{tabular}
\end{table}

\emph{The comparison spans three tiers---no evaluation (Self-Reflection), single-turn QA evaluation~\citep{liu2026skillforge}, and dynamic interaction evaluation (SkillEvo)---each run for four rounds. Other single-turn paradigms, including SkillOpt~\citep{yang2026skillopt}, EvoSkill~\citep{alzubi2026evoskill}, Trace2Skill~\citep{ni2026trace2skill}, SkillCAT~\citep{chen2026skillcat}, GEPA~\citep{agrawal2026gepa}, and TextGrad~\citep{yuksekgonul2024textgrad}, all target automatically verifiable single-turn tasks and cannot accommodate the layer-by-layer intent exposure and dialogue-level judgment of multi-turn consultation; we therefore adopt~\citep{liu2026skillforge} as the representative of that tier. Section~ holds attribution, revision, and governance fixed and replaces only the feedback source, isolating the effect of multi-turn feedback.}

\subsubsection{Evaluation metrics}
\label{sec:metrics}

\begin{itemize}[leftmargin=*,itemsep=1pt]
\item \textbf{Overall TSR}: the fraction of tasks solved on the evaluation set. The Verifier assigns each ticket a continuous knowledge score in $[0,100]$, and the ticket counts as solved if the score reaches the passing threshold of 60 \emph{and} no key condition of the task definition is missing.
\item \textbf{Exposed-intent response accuracy} ($s_C$): the agent's weighted accuracy on intents already raised during simulation; distinct from Overall TSR.
\item \textbf{Intent coverage} ($c_U$): the fraction of key intents raised, $c_U = |\mathcal{K}_{\text{asked}}| / |\mathcal{K}|$.
\item \textbf{Cross-round regression rate} (RegR): the fraction of tickets that passed in the previous round but fail in the current one, quantifying the damage multi-round revision inflicts on existing capability,
\begin{equation}
RegR(r) = \frac{|\{\,t : s_{r-1}(t) \ge 60 \;\wedge\; s_{r}(t) < 60\,\}|}{|\{\,t : s_{r-1}(t) \ge 60\,\}|}.
\end{equation}
A lower RegR indicates a more stable revision, and directly measures the fact-consistency constraint.
\item \textbf{Knowledge bloat} (Bloat): the growth in the total line count of \texttt{SKILL.md} and all reference files relative to the production baseline $S_0$,
\begin{equation}
Bloat(S_t) = \frac{\text{lines}(S_t) - \text{lines}(S_0)}{\text{lines}(S_0)}.
\end{equation}
A lower Bloat indicates a more concise revision, and directly measures the structural-consistency constraint.
\end{itemize}

\textbf{Verifier reliability.} To validate the Verifier as a proxy for human judgment, we sample a random subset of the evaluation set for independent labeling by domain experts. Agreement between the Verifier's verdicts and the human consensus exceeds 90\%, indicating that automatic verdicts form a reliable basis for driving evolution feedback.

\subsection{Main results}

\begin{table}[ht]
\centering
\small
\caption{Per-round TSR of each method on the evaluation set (\%).}
\label{tab:main}
\begin{tabular}{lccccc}
\toprule
\textbf{Method} & \textbf{Init} & \textbf{R1} & \textbf{R2} & \textbf{R3} & \textbf{R4} \\
\midrule
Original Skill & 30.0 & --- & --- & --- & --- \\
Self-Reflection & 30.0 & 59.2 & 58.7 & 57.4 & 58.8 \\
Single-turn QA & 30.0 & 58.9 & 64.5 & 65.7 & 66.4 \\
\textbf{SkillEvo} & 30.0 & 59.4 & 71.3 & 77.9 & \textbf{81.8} \\
\bottomrule
\end{tabular}
\end{table}

\emph{R1--R4 report per-round TSR. For the methods that possess evaluation feedback (Single-turn QA and SkillEvo), the version reported at each round is the best one selected on the development set up to that round, which accounts for the monotone non-decreasing trend; Self-Reflection has no feedback and therefore cannot select, so its current-round version is reported and oscillates.}

The differences among the three methods reduce to a single question: \textbf{can the feedback keep supplying an evolution gradient}---that is, does a new failure signal capable of guiding the next round survive each revision?

Self-Reflection possesses no gradient in the absence of evaluation, so multi-round blind editing merely oscillates around its first-round level. Single-turn QA supplies a first-round gradient, but a single question--answer pair reaches only the gaps present in the user's opening statement; once those are patched the gradient decays, and TSR climbs from 58.9 to no more than 66.4 with sharply diminishing marginal gains. The two differ in one respect that matters: Single-turn QA, decaying gradient notwithstanding, retains an evaluation gate that intercepts degradation, so its curve plateaus rather than falls, whereas Self-Reflection has neither gradient nor gate and oscillates without substantive evolution.

The multi-turn interaction of SkillEvo changes how gradients arise. Follow-up questions and clarification expose defects layer by layer: knowledge patched in the current round lets the dialogue proceed further and reach the next layer of defects, previously masked by shallower failures. Each round of revision therefore not only consumes gradients but generates new ones, and TSR rises steadily from 59.4 to 81.8. Single-turn evaluation cannot replicate this self-renewing behavior, because its failure surface is observed in full during the first round, whereas the multi-turn failure surface keeps unfolding as Skill capability improves.

\subsection{Ablation study}
\label{sec:ablation}

\begin{table}[ht]
\centering
\small
\caption{Ablation (evaluation-set TSR, \%).}
\label{tab:ablation}
\begin{tabular}{lc}
\toprule
\textbf{Variant} & \textbf{Overall TSR} \\
\midrule
SkillEvo (Full) & 81.8 \\
(a) Single-turn QA & 66.4 \\
(b) w/o Governance & 78.6 \\
\bottomrule
\end{tabular}
\end{table}

Variant (a) replaces multi-turn interaction with single-turn QA evaluation while leaving attribution, revision, and governance untouched, rendering it substantively equivalent to the Single-turn QA baseline of the main experiment. Both reach 66.4 at R4: removing multi-turn interaction closes the gap entirely, and SkillEvo's 15.4-point lead is therefore attributable to the feedback source itself.

Variant (b) removes the governance layer while leaving feedback, attribution, and revision untouched; TSR falls to 78.6 ($-3.2$), a far smaller drop than under the feedback-source ablation. The value of governance lies not in raising the score but in preventing degradation from accumulating across rounds (Table~5).

\subsection{Trustworthy feedback and controllable governance}

We now examine the two pillars of the framework in turn: whether the evolution gradient is trustworthy, and whether its direction is controllable.

\subsubsection{Trustworthy feedback: dual-sided orthogonal evaluation}

Trustworthy feedback presupposes a reliable simulator side: should the simulated user fail to raise the real intents, an agent failure cannot be separated into insufficient capability and misjudgment induced by a distorted simulation. Since attributability is already secured by the three-way classification, we focus here on the three measurable properties of coverage, fidelity, and accuracy.

\begin{table}[ht]
\centering
\small
\caption{Dual-sided orthogonal evaluation (\%).}
\label{tab:trust}
\begin{tabular}{llc}
\toprule
\textbf{Side / property} & \textbf{Metric} & \textbf{Value} \\
\midrule
Simulator, coverage & $c_U$ (intent coverage) & 98.9 \\
Simulator, fidelity & $\rho$ (human-rated similarity) & 95.3 \\
Agent, accuracy & $s_C$ (exposed-intent accuracy) & 71.1 \\
\bottomrule
\end{tabular}
\end{table}

Coverage reaches 98.9\%, indicating that key intents are raised almost exhaustively; the remaining 1\% of low-coverage samples are isolated from the revision loop. To exclude the possibility that intent-state-machine gating inflates this figure artificially, we add an external validation: 200 simulated dialogues are sampled from the evaluation set across all 9 Skills, and two domain experts blindly compare each against its real ticket along intent expression, information-reveal pace, and emotion trajectory. Agreement across the three dimensions reaches 95.3\%, confirming that the simulator raises intents in a manner close to a real user rather than mechanically walking a checklist.

Under this premise, the agent answers exposed intents with 71.1\% accuracy. The value reflects the Skill's knowledge-coverage gaps and is the direct source of the evolution gradient; it is not a whole-dialogue resolution rate. The two sides are orthogonal by construction, which is what prevents unexposed or distortedly exposed capability from being misread as agent failure.

\subsubsection{Controllable governance: regression and bloat}
\label{sec:governance}

The governance layer earns its place by preventing degradation, not by raising scores. Beyond TSR, we therefore measure directly how much multi-round revision damages existing capability and knowledge structure, using RegR and Bloat.

\begin{table}[ht]
\centering
\small
\caption{Cross-round regression rate RegR (\%).}
\label{tab:regr}
\begin{tabular}{cccc}
\toprule
\textbf{R1$\to$2} & \textbf{R2$\to$3} & \textbf{R3$\to$4} & \textbf{First-to-last change} \\
\midrule
28.2 & 24.4 & 21.1 & $-7.1$ \\
\bottomrule
\end{tabular}
\end{table}

\begin{table}[ht]
\centering
\small
\caption{Knowledge bloat (cumulative growth relative to $S_0$, \%).}
\label{tab:bloat}
\begin{tabular}{lcp{5.6cm}}
\toprule
\textbf{Setting} & \textbf{Cumulative bloat} & \textbf{Remark} \\
\midrule
With governance (SkillEvo Full) & $+2.8$ & Growth concentrates in the first round and then tapers off \\
Without governance & $+16.2$ & Bloat accumulates round after round with no dissolution mechanism \\
\bottomrule
\end{tabular}
\end{table}

RegR declines across the three transitions, indicating that the governance layer constrains cross-round regression. The comparison on Bloat is more direct still: cumulative growth reaches only 2.8\% under governance, stabilizing after the first round of knowledge supplementation, against 16.2\% without it---nearly six times larger---which shows the structural-consistency constraint to be effective against knowledge bloat. That TSR improves by 51.8 points while volume barely changes indicates that the capability gain arises from revising existing knowledge correctly rather than from expanding the text.

\section{Conclusion}

We have presented SkillEvo, which reconstructs skill self-evolution along two dimensions: trustworthy feedback and controllable governance. Our core insight is that the bottleneck of sustained skill evolution lies neither in editing capability nor in the number of iterations, but in whether evaluation feedback keeps supplying trustworthy evolution gradients. On 9 production Skills, SkillEvo improves TSR by 51.8 points over the original Skills and by 15.4 points over single-turn-QA-driven evolution; our ablation attributes this lead to multi-turn interaction feedback itself, and the trustworthiness evaluation confirms the reliability of the simulator side in both coverage and fidelity. Reliable skill evolution thus rests on two requirements: a high-quality feedback loop that exposes interaction defects and isolates simulation distortion, and a governance mechanism that actively maintains the Skill as a structured knowledge system.

\clearpage

\section*{Ethics Statement}

Our study is conducted on tickets escalated to human agents in a production customer-support system, and we have taken the following measures to ensure responsible data use. All tickets are de-identified before entering the pipeline: account identifiers, order numbers, phone numbers, personal names, resource and instance identifiers, and temporary signed links are removed, and the signal-extraction prompt explicitly forbids extracting such case-specific details into a Skill. What the framework consolidates is therefore stable, non-personal product knowledge---billing rules, console navigation paths, product behaviors, and official documentation links---rather than any information about individual users. The data are used under the platform's terms of service and internal data-governance policy, and the study involves no human-subject experimentation beyond the expert annotation, for which annotators were members of the support engineering team working within their normal duties.

We also note a risk intrinsic to any framework that writes knowledge back automatically: an erroneous revision, once merged, would be served to real users. SkillEvo mitigates this in two ways. The governance layer treats fact consistency as a hard constraint and rejects any candidate that deletes stable facts from the production baseline, and no revision reaches production without human confirmation---the loop terminates at a reviewed candidate rather than at an automatic rollout. We regard this human checkpoint as a necessary condition for deploying self-evolving knowledge in user-facing systems, not an optional safeguard.

\section*{Reproducibility Statement}

We report the components required to reimplement SkillEvo. Appendix~B gives the core constraints of every prompt in the pipeline---user simulation, scenario synthesis, signal extraction, signal merging, attribution, the three editor modes, quality inspection, and governance---together with the two scoring rubrics that define the Verifier's verdicts. Appendix~C states the full evolution loop as pseudocode, Appendix~F lists every hyperparameter with its value, Appendix~E specifies the model assignment and the tool allow-list of the evaluation environment, and Appendix~G describes the end-to-end pipeline and the two-level loop structure. Appendix~D traces one ticket end to end---scenario, failed trajectory, Verifier verdict, attribution output, revision diff, and the interaction after the update---so that each stage can be checked against a concrete instance.

Two limitations bear on external reproduction. First, the tickets originate from a production support system and carry both user-privacy and commercial-confidentiality constraints, so the dataset cannot be released; the method, however, is independent of this particular ticket source and applies to any setting in which multi-turn consultation logs with human reference solutions are available. Second, on model choice, the only architectural requirement is Generator $\neq$ Evaluator: the specific models we use may be substituted by any two models from different families without altering the framework, since no component depends on a model-specific capability.

% ...

\bibliographystyle{iclr2026_conference}
\bibliography{SkillEvo}

@article{wang2024survey,
  title={A survey on large language model based autonomous agents},
  author={Wang, Lei and Ma, Chen and Feng, Xueyang and Zhang, Zeyu and Yang, Hao and Zhang, Jingsen and Chen, Zhiyuan and Tang, Jiakai and Chen, Xu and Lin, Yankai and others},
  journal={Frontiers of computer science},
  volume={18},
  number={6},
  pages={186345},
  year={2024},
  publisher={Springer}
}

@inproceedings{chen2024automatic,
  title={Automatic root cause analysis via large language models for cloud incidents},
  author={Chen, Yinfang and Xie, Huaibing and Ma, Minghua and Kang, Yu and Gao, Xin and Shi, Liu and Cao, Yunjie and Gao, Xuedong and Fan, Hao and Wen, Ming and others},
  booktitle={Proceedings of the Nineteenth European Conference on Computer Systems},
  pages={674--688},
  year={2024}
}

@article{zhou2023d,
  title={D-bot: Database diagnosis system using large language models},
  author={Zhou, Xuanhe and Li, Guoliang and Sun, Zhaoyan and Liu, Zhiyuan and Chen, Weize and Wu, Jianming and Liu, Jiesi and Feng, Ruohang and Zeng, Guoyang},
  journal={arXiv preprint arXiv:2312.01454},
  year={2023}
}

@inproceedings{zhao2025agent,
  title={Agent-in-the-Loop: A Data Flywheel for Continuous Improvement in LLM-based Customer Support},
  author={Zhao, Cen and Zhang, Tiantian and Su, Hanchen and Zhang, Yufeng and Su, Shaowei and Xu, Mingzhi and Liu, Yu and Han, Wei and Werner, Jeremy and Cheng, Claire Na and others},
  booktitle={Proceedings of the 2025 conference on empirical methods in natural language processing: Industry track},
  pages={1919--1930},
  year={2025}
}

@article{liu2026skillforge,
  title={Skillforge: Forging domain-specific, self-evolving agent skills in cloud technical support},
  author={Liu, Xingyan and Luo, Xiyue and Li, Linyu and Huang, Ganghong and Liu, Jianfeng and Qiao, Honglin},
  journal={arXiv preprint arXiv:2604.08618},
  year={2026}
}

@article{wei2022chain,
  title={Chain-of-thought prompting elicits reasoning in large language models},
  author={Wei, Jason and Wang, Xuezhi and Schuurmans, Dale and Bosma, Maarten and Xia, Fei and Chi, Ed and Le, Quoc V and Zhou, Denny and others},
  journal={Advances in neural information processing systems},
  volume={35},
  pages={24824--24837},
  year={2022}
}

@article{yao2023tree,
  title={Tree of thoughts: Deliberate problem solving with large language models},
  author={Yao, Shunyu and Yu, Dian and Zhao, Jeffrey and Shafran, Izhak and Griffiths, Tom and Cao, Yuan and Narasimhan, Karthik},
  journal={Advances in neural information processing systems},
  volume={36},
  pages={11809--11822},
  year={2023}
}

@article{yao2024tau,
  title={{$\tau$-bench}: A Benchmark for Tool-Agent-User Interaction in Real-World Domains},
  author={Yao, Shunyu and Shinn, Noah and Razavi, Pedram and Narasimhan, Karthik},
  journal={arXiv preprint arXiv:2406.12045},
  year={2024}
}

@inproceedings{wang2025ecom,
  title={Ecom-bench: Can llm agent resolve real-world e-commerce customer support issues?},
  author={Wang, Haoxin and Peng, Xianhan and Cheng, Huang and Huang, Yizhe and Gong, Ming and Yang, Chenghan and Liu, Yang and Lin, Jiang},
  booktitle={Proceedings of the 2025 Conference on Empirical Methods in Natural Language Processing: Industry Track},
  pages={276--284},
  year={2025}
}

@article{xu2025voiceagenteval,
  title={VoiceAgentEval: A Dual-Dimensional Benchmark for Expert-Level Intelligent Voice-Agent Evaluation of Xbench's Professional-Aligned Series},
  author={Xu, Pengyu and Li, Shijia and Sun, Ao and Zhang, Feng and Li, Yahan and Wu, Bo and Ma, Zhanyu and Li, Jiguo and Xu, Jun and Gao, Jiuchong and others},
  journal={arXiv preprint arXiv:2510.21244},
  year={2025}
}

@inproceedings{shea2026sage,
  title={SAGE: A Top-Down Bottom-Up Knowledge-Grounded User Simulator for Multi-turn AGent Evaluation},
  author={Shea, Ryan and Lu, Yunan and Qiu, Liang and Yu, Zhou},
  booktitle={Findings of the Association for Computational Linguistics: EACL 2026},
  pages={2816--2839},
  year={2026}
}

@article{madaan2023self,
  title={Self-refine: Iterative refinement with self-feedback},
  author={Madaan, Aman and Tandon, Niket and Gupta, Prakhar and Hallinan, Skyler and Gao, Luyu and Wiegreffe, Sarah and Alon, Uri and Dziri, Nouha and Prabhumoye, Shrimai and Yang, Yiming and others},
  journal={Advances in neural information processing systems},
  volume={36},
  pages={46534--46594},
  year={2023}
}

@article{yang2026skillopt,
  title={Skillopt: Executive strategy for self-evolving agent skills},
  author={Yang, Yifan and Gong, Ziyang and Huang, Weiquan and Yang, Qihao and Zhou, Ziwei and Huang, Zisu and Li, Yan and Gao, Xuemei and Dai, Qi and Liu, Bei and others},
  journal={arXiv preprint arXiv:2605.23904},
  year={2026}
}

@article{zhang2025agent,
  title={Which agent causes task failures and when? on automated failure attribution of llm multi-agent systems},
  author={Zhang, Shaokun and Yin, Ming and Zhang, Jieyu and Liu, Jiale and Han, Zhiguang and Zhang, Jingyang and Li, Beibin and Wang, Chi and Wang, Huazheng and Chen, Yiran and others},
  journal={arXiv preprint arXiv:2505.00212},
  year={2025}
}

@article{alzubi2026evoskill,
  title={Evoskill: Automated skill discovery for multi-agent systems},
  author={Alzubi, Salaheddin and Provenzano, Noah and Bingham, Jaydon and Chen, Weiyuan and Vu, Tu},
  journal={arXiv preprint arXiv:2603.02766},
  year={2026}
}

@article{ni2026trace2skill,
  title={Trace2skill: Distill trajectory-local lessons into transferable agent skills},
  author={Ni, Jingwei and Liu, Yihao and Liu, Xinpeng and Sun, Yutao and Zhou, Mengyu and Cheng, Pengyu and Wang, Dexin and Zhao, Erchao and Jiang, Xiaoxi and Jiang, Guanjun},
  journal={arXiv preprint arXiv:2603.25158},
  year={2026}
}

@article{chen2026skillcat,
  title={Skillcat: Contrastive assessment and topology-aware skill self-evolution for llm agents},
  author={Chen, Kunfeng and Zhong, Qihuang and Liu, Juhua and Du, Bo},
  journal={arXiv preprint arXiv:2606.13317},
  year={2026}
}

@inproceedings{agrawal2026gepa,
  title={Gepa: Reflective prompt evolution can outperform reinforcement learning},
  author={Agrawal, Lakshya A and Tan, Shangyin and Soylu, Dilara and Ziems, Noah and Khare, Rishi and Opsahl-Ong, Krista and Singhvi, Arnav and Shandilya, Herumb and Ryan, Michael J and Jiang, Meng and others},
  booktitle={International Conference on Learning Representations},
  volume={2026},
  pages={8479--8565},
  year={2026}
}

@article{yuksekgonul2024textgrad,
  title={Textgrad: Automatic" differentiation" via text},
  author={Yuksekgonul, Mert and Bianchi, Federico and Boen, Joseph and Liu, Sheng and Huang, Zhi and Guestrin, Carlos and Zou, James},
  journal={arXiv preprint arXiv:2406.07496},
  year={2024}
}

@misc{zhang2025agentskills,
  title        = {Equipping Agents for the Real World with Agent Skills},
  author       = {Zhang, Barry and Lazuka, Keith and Murag, Mahesh},
  howpublished = {Anthropic Engineering Blog},
  year         = {2025}
}

@article{li2026skillsbench,
  title={SkillsBench: Benchmarking how well agent skills work across diverse tasks},
  author={Li, Xiangyi and Liu, Yimin and Chen, Wenbo and You, Bingran and Di, Zonglin and He, Yifeng and Zheng, Shenghan and Choe, Kyoung Whan and Sun, Jiankai and Wang, Shuyi and others},
  journal={arXiv preprint arXiv:2602.12670},
  year={2026}
}

@article{jiang2026sok,
  title={SoK: Agentic Skills--Beyond Tool Use in LLM Agents},
  author={Jiang, Yanna and Li, Delong and Deng, Haiyu and Ma, Baihe and Wang, Xu and Wang, Qin and Yu, Guangsheng},
  journal={arXiv preprint arXiv:2602.20867},
  year={2026}
}

@article{shen2026skillfoundry,
  title={Skillfoundry: Building self-evolving agent skill libraries from heterogeneous scientific resources},
  author={Shen, Shuaike and Cheng, Wenduo and Ma, Mingqian and Turcan, Alistair and Zhang, Martin Jinye and Ma, Jian},
  journal={arXiv preprint arXiv:2604.03964},
  year={2026}
}

@article{wang2026skillx,
  title={Skillx: Automatically constructing skill knowledge bases for agents},
  author={Wang, Chenxi and Yu, Zhuoyun and Xie, Xin and Yao, Wuguannan and Fang, Runnan and Qiao, Shuofei and Cao, Kexin and Zheng, Guozhou and Qi, Xiang and Zhang, Peng and others},
  journal={arXiv preprint arXiv:2604.04804},
  year={2026}
}

@article{zhang2026coevoskills,
  title={Coevoskills: Self-evolving agent skills via co-evolutionary verification},
  author={Zhang, Hanrong and Fan, Shicheng and Zou, Henry Peng and Chen, Yankai and Wang, Zhenting and Zhou, Jiayu and Li, Chengze and Huang, Wei-Chieh and Yao, Yifei and Zheng, Kening and others},
  journal={arXiv preprint arXiv:2604.01687},
  year={2026}
}

@article{li2026organizing,
  title={Organizing, orchestrating, and benchmarking agent skills at ecosystem scale},
  author={Li, Hao and Mu, Chunjiang and Chen, Jianhao and Ren, Siyue and Cui, Zhiyao and Zhang, Yiqun and Bai, Lei and Hu, Shuyue},
  journal={arXiv preprint arXiv:2603.02176},
  year={2026}
}

@article{xie2026steve,
  title={Steve-Evolving: Open-World Embodied Self-Evolution via Fine-Grained Diagnosis and Dual-Track Knowledge Distillation},
  author={Xie, Zhengwei and Chen, Zhisheng and Weng, Ziyan and Wu, Tingyu and Li, Chenglong and Zhang, Vireo and Wang, Kun},
  journal={arXiv e-prints},
  pages={arXiv--2603},
  year={2026}
}

@article{ouyang2026skillos,
  title={Skillos: Learning skill curation for self-evolving agents},
  author={Ouyang, Siru and Yan, Jun and Chen, Yanfei and Han, Rujun and Wang, Zifeng and Mishra, Bhavana Dalvi and Meng, Rui and Li, Chun-Liang and Jiao, Yizhu and Zha, Kaiwen and others},
  journal={arXiv preprint arXiv:2605.06614},
  year={2026}
}

@article{huang2025audited,
  title={Audited skill-graph self-improvement for agentic llms via verifiable rewards, experience synthesis, and continual memory},
  author={Huang, Ken and Huang, Jerry},
  journal={arXiv preprint arXiv:2512.23760},
  year={2025}
}

@inproceedings{dai2026sead,
  title={Sead: Self-evolving agent for multi-turn service dialogue},
  author={Dai, Yuqin and Gao, Ning and Zhang, Wei and Wang, Jie and Wu, Ruiyuan and Wang, Jinpeng and Wang, Chaozheng and others},
  booktitle={Findings of the Association for Computational Linguistics: ACL 2026},
  pages={3674--3684},
  year={2026}
}

@article{gao2025survey,
  title={A survey of self-evolving agents: What, when, how, and where to evolve on the path to artificial super intelligence},
  author={Gao, Huan-ang and Geng, Jiayi and Hua, Wenyue and Hu, Mengkang and Juan, Xinzhe and Liu, Hongzhang and Liu, Shilong and Qiu, Jiahao and Qi, Xuan and Wu, Yiran and others},
  journal={arXiv preprint arXiv:2507.21046},
  year={2025}
}

@inproceedings{wang2024rcagent,
  title={Rcagent: Cloud root cause analysis by autonomous agents with tool-augmented large language models},
  author={Wang, Zefan and Liu, Zichuan and Zhang, Yingying and Zhong, Aoxiao and Wang, Jihong and Yin, Fengbin and Fan, Lunting and Wu, Lingfei and Wen, Qingsong},
  booktitle={Proceedings of the 33rd ACM international conference on information and knowledge management},
  pages={4966--4974},
  year={2024}
}

\appendix
\section*{\Huge\bfseries Appendix}
\vspace{0.6em}

\section{Evaluation scenarios and dataset}
\label{app:dataset}

\begin{table}[H]
\centering
\small
\caption{Evaluation scenarios and dataset.}
\label{tab:dataset}
\begin{tabular}{p{3.1cm}p{4.3cm}p{4.0cm}r}
\toprule
\textbf{Scenario category} & \textbf{Skill} & \textbf{Cloud service} & \textbf{\#Tickets} \\
\midrule
Marketing & \texttt{dianshi-consultation} & Dianshi campaign platform & 400 \\
Dev \& collaboration tools & \texttt{code-assistant-consultation} & CodeBuddy coding assistant & 400 \\
Storage & \texttt{cos-consultation} & Cloud Object Storage (COS) & 200 \\
Storage & \texttt{cbs-consultation} & Cloud Block Storage (CBS) & 160 \\
Dev \& collaboration tools & \texttt{cloudbase-consultation} & CloudBase & 200 \\
AI \& LLM platforms & \texttt{tokenhub-consultation} & TokenHub LLM service platform & 200 \\
AI \& LLM platforms & \texttt{tencent-adp-consultation} & Agent Development Platform (ADP) & 200 \\
Networking \& edge & \texttt{edgeone-consultation} & EdgeOne edge security acceleration & 200 \\
Compute & \texttt{scf-consultation} & Serverless Cloud Function (SCF) & 40 \\
\midrule
\textbf{Total} & \textbf{9 Skills} & \textbf{6 categories} & \textbf{2{,}000} \\
\bottomrule
\end{tabular}
\end{table}

\section{Prompt collection}
\label{app:prompts}

\subsection{General prompts}
\label{app:prompts_general}

\emph{Note. Each prompt below is a condensation of the core constraints of the full prompt in production, retaining the behavioral constraints, output format, and adjudication rules while omitting repetitive exemplars and format-validation directives.}

\subsubsection{User simulation prompt}
\label{app:usersim}

The system prompt of the simulated user lays down its behavioral rules: (1) disclose information progressively, opening with a single sentence stating the main complaint; (2) send fragmented messages, splitting content into short messages as a real person would; (3) cooperate but stay in role, getting stuck and asking for help when the agent requests complex operations; (4) let emotion evolve with progress; (5) advance one concrete intent per turn; and (6) never initiate interactions beyond the agent's capability. Each decision emits an action block (\texttt{\textless reason\textgreater} + \texttt{\textless agenda\_check\textgreater} + \texttt{\textless action\textgreater} + \texttt{\textless say\textgreater}), in which the \texttt{agenda\_check} tag reports the topics raised during the current turn so that the intent state machine can update its state.

\begin{verbatim}
You are a real cloud-platform user asking online support for help.
You are not an AI; never reveal any AI/assistant identity.

## What you will / will not do

You WILL:
- Describe the problem you hit (malfunction, billing question, service
  unavailable, error message).
- Supply information on request: account ID, order ID, phone number,
  instance ID, error code, screenshot content.
- Restate, from the behavior facts, the operations you performed and the
  symptoms you saw, but never re-execute operations mid-dialogue.
- Confirm resolution: say "it works now" if the solution is effective;
  keep asking if it is not.

You will NOT:
- Read logs, capture packets, use dev tools, or SSH into a server. If asked,
  say "I don't know how" / "please check it for me".
- Diagnose root causes yourself; only describe symptoms ("it won't open",
  "the credit never arrived", "it throws an error").
- Fabricate information. If asked for an order/account ID, either use the
  scenario information or say "let me look for it".

## Behavioral rules

1. Disclose progressively: open with one sentence stating the main
   complaint; add one detail only when asked for it.
2. Send fragmented messages: sometimes split one utterance into two or
   three short messages, as a real person would.
3. Cooperate but stay in role: answer questions about information; get
   stuck and ask for help when asked to perform complex operations.
4. Let emotion follow progress: calm down on progress, grow impatient
   when going in circles or waiting too long.
5. Advance one concrete intent per turn: pure pleasantries, pure urging,
   or pure emotion may not constitute a turn on their own.
6. Never initiate interactions beyond the agent's capability: do not
   request human handoff, back-office lookups, or proxy operations.

## Intent check (at every decision)

In the <agenda_check> tag, report which topics you raised this turn,
listing the verbatim topics from the agenda. The state machine updates
intent status from your report; the dialogue may end only after all
intents have been raised.

## Output format

Emit exactly one action block per turn, wrapped in XML tags:
<reason>...</reason>
<agenda_check>topic 1 (newline) topic 2</agenda_check>
<action>send_text | done</action>
<say>the user's utterance</say>
\end{verbatim}

\subsubsection{Scenario synthesis prompt}
\label{app:scenario}

The scenario-synthesis prompt constructs a hidden task from the complete dialogue trace of a ticket, comprising \texttt{opening\_message}, \texttt{behavior\_facts}, \texttt{emotion\_trajectory}, and \texttt{target\_keywords} with priorities, the last of which forms the intent agenda. The \texttt{expected\_solution} is passed separately to the Verifier as the human reference and is never injected into the User Agent.

\begin{verbatim}
You are reconstructing the user scenario of a real ticket for simulated-user
evaluation, producing the persona for the simulated user agent together with
the reference solution for evaluation.

## Requirements

### 1. opening_message
In a real user's voice, colloquially and briefly summarize the core problem
the user initially wants solved. State only the perceived symptom and the
request; never reveal the technical root cause.

### 2. emotion_trajectory
The user's emotional state and possible points of friction, as one sentence
or a short arrow chain, e.g. "confused, doubts the refund amount" /
"impatient -> urging -> dissatisfied" / "patient and cooperative".

### 3. behavior_facts
What the user did before the consultation and what result/symptom was seen.
Include only: the user's own operations, observed symptoms, account and
environment information. Exclude: agent behavior, human handoff, user
intent, user emotion, and any in-dialogue interaction.

### 4. target_keywords
A key intent is the ultimate purpose of the consultation---answering it
alone would let the user leave satisfied. A minor intent is a supplement,
follow-up, or extension around a key intent; like key intents, all minor
intents must be raised before the dialogue may end. Different stages of the
same request merge into one key intent; two requests get separate key
intents only if they are fully independent.

### 5. expected_solution
Extract only knowledge content learnable by a Skill (rule statements,
operation paths, constraints, troubleshooting steps). Exclude proxy
operations and back-office lookups. At most 200 words.

## No answer leakage
- opening_message and target_keywords must be requests from the user's
  perspective; never include the human agent's operation steps, solution,
  or diagnostic conclusion.
- expected_solution is for evaluation only and is never injected into the
  simulated-user prompt.

## Output JSON
{"opening_message": "...", "emotion_trajectory": "...",
"behavior_facts": "...",
"target_keywords": [{"topic": "...", "priority": "key"}],
"expected_solution": "..."}
\end{verbatim}

\subsubsection{Signal extraction prompts}
\label{app:signal}

Signal extraction proceeds in two stages: first adjudicate whether the human agent genuinely resolved the problem and whether the experience is reusable (\texttt{judge\_outcome}); then extract a structured learning signal from the tickets that pass (\texttt{extract\_signal}).

\textbf{judge\_outcome prompt.}

\begin{verbatim}
Context: you are distilling reusable factual business knowledge from
human-escalated cloud-product tickets so that the AI can answer customer
questions directly next time. Adjudicate the human handling outcome of this
ticket. Return only a JSON object with fields resolved, reusable_by_ai,
and human_action_type.

Field definitions:
- resolved: bool. Did the human agent genuinely resolve the user's problem
  (explicit thanks / confirmation / symptom disappearance / an actionable
  solution was given)?
- reusable_by_ai: bool. Can the human experience be reused directly by the
  AI next time? It must not be a merely procedural action ("escalate /
  gather more / file a ticket"), nor a case-specific investigation that
  depends on diagnostic tooling.
- human_action_type: enum, one of resolved / handoff / waiting_for_user /
  internal_operation / document_guidance / configuration_guidance / unknown.

Adjudication rules:
- "Escalate / please elaborate / file a ticket" alone does not count as
  resolved; reusable_by_ai=false.
- If the human agent supplied a concrete business rule, a feature location
  or console navigation path, an explanation of product behavior, or a
  console/documentation link, then reusable_by_ai=true.
- For incident tickets where the human only investigated, localized, or
  repaired without giving the user an actionable step, reusable_by_ai=false.
- For incident tickets where the human gave the user a lightweight step
  executable without cloud-support or back-office involvement (run as
  administrator, clear local cache or DNS, edit a local config file, adjust
  browser or client settings), reusable_by_ai=true.
- is_badcase=true means the AI already attempted and failed; adjudicate
  nonetheless by the final human outcome.

Ticket: {ticket_id}
is_badcase: {is_badcase}
Session material:
{session_material}
\end{verbatim}

\textbf{extract\_signal prompt} (core constraints).

\begin{verbatim}
Context: you are distilling reusable factual business knowledge from
human-escalated cloud-product tickets so that the AI can answer customer
questions directly next time. This prompt covers both consultation and
incident tickets; knowledge_type drives downstream filtering. Extract a
reusable learning signal from the de-identified ticket and return only a
JSON object with fields skill_topic, user_problem, knowledge_facts,
suggested_change, generalizable, knowledge_type, and quality_score.

Learning objective: distill factual business knowledge (product behavior,
rule conditions, feature locations, status explanations, pricing and
entitlements) together with lightweight troubleshooting steps the user can
perform. Such knowledge must be directly reusable by the AI without human
involvement or back-office tooling.

Requirements:
- Except for knowledge_facts, which is an array of strings, no field may
  return an array or a nested object.
- Contain no sensitive information; if the content cannot be generalized,
  set generalizable=false.
- Where the human reply contains product-behavior explanations, rule
  conditions, documentation links, or lightweight user-executable
  troubleshooting steps, that is high-value knowledge and must be extracted
  into knowledge_facts in full.
- Mandatory filtering:
  * Tickets containing only escalation / ticket filing / an operation card /
    "please elaborate", with no human resolution knowledge
    -> generalizable=false.
  * user_problem must state the real business problem abstracted into a
    category; never write "the user asked for a human", and never include a
    concrete resource ID, instance name, or ticket number.
- knowledge_facts:
  * An array of strings holding reusable factual knowledge extracted from
    the human reply.
  * Each entry must be self-contained, covering business or product
    behavior, rule conditions, feature location or console navigation path,
    officially stable pricing / entitlement / link, and lightweight
    user-executable troubleshooting steps.
  * Never extract case-specific details: a particular user's refund or
    compensation amount, resource ID, ticket number, user name, a versioned
    temporary download link, or a case-specific investigation path.
  * Never write procedural steps or methodological guidance.
  * If the ticket contains no reusable factual knowledge, return [].
- knowledge_type:
  * consultation: the user is asking about product rules, behavior,
    conditions, paths, status, or differences, and the human answered with
    an explanation, a document, or console guidance.
  * light_troubleshooting: the user hit a fault or error, but the human
    solution is a lightweight operation the user can perform.
  * heavy_incident: the fault requires cloud-support involvement,
    back-office investigation, internal operations, or diagnostic tooling;
    generalizable=false.
- quality_score: integer 1-10, rating the ticket's value as an evaluation
  sample (see the rubric in the scoring-rubric appendix).

Classification: {classification}
Ticket: {ticket_id}
is_badcase: {is_badcase}
Human outcome: resolved={outcome_resolved}, action={outcome_action}
Session material:
{session_material}
\end{verbatim}

\subsubsection{Signal merging prompt}
\label{app:merging}

The signal merger consolidates multiple learning signals within the same topic bucket into one canonical signal, deduplicating \texttt{knowledge\_facts} semantically.

\begin{verbatim}
You are a signal merger. Below are several learning signals from the same
topic bucket (drawn from different tickets but sharing a topic). Your task
is to merge them into one distilled signal, deduplicating knowledge_facts
semantically.

## Merging criteria

1. knowledge_facts: take the union, then deduplicate semantically.
   - Multiple phrasings of one fact merge into a single canonical
     statement.
   - Retain every concrete fact (version number / path / value / rule /
     link); lose no information.
   - Discard vague, duplicated, or uninformative statements.
2. user_problem: take the entry that best summarizes the merged knowledge;
   if several emphasize different aspects, concatenate into one sentence.
3. skill_topic: take the most general entry.
4. suggested_change: take the most concrete and actionable entry; if
   entries are complementary, concatenate without information loss.

## Output requirements

Return exactly one JSON object with four fields:
- skill_topic: str
- user_problem: str
- knowledge_facts: list[str]
- suggested_change: str

Emit no Markdown code fence and no explanatory prose.

## Input (JSON array)

{bucket_json}
\end{verbatim}

\subsubsection{Attributor prompt}
\label{app:attributor}

The attribution prompt directs the Attributor to compare the human handling process, the simulated dialogue, and the Verifier's evidence, to assign the failure to one of Knowledge Gap, Capability Limit, or Evaluation Noise, and to output \texttt{knowledge\_facts} together with \texttt{evidence}.

\begin{verbatim}
Context: diagnose the root cause of an unresolved (or low-scoring) ticket in
a customer-support skill evaluation. You will see two dialogues for the same
ticket:
1. [Human dialogue]: the full record of a human agent handling the ticket
   (the correct reference).
2. [Simulated dialogue]: the bot handling the same problem (the subject
   under evaluation).

Your task: contrast the two, identify where the bot falls short of the human
agent, determine the root-cause type, and produce a structured signal that
can be used directly to edit the skill.

## Root-cause taxonomy (3 classes; the crux is "who can fix it")

- knowledge_gap: missing / stale / mis-routed knowledge. The human agent
  supplied a factual item (version number, feature location, console path,
  billing rule, product behavior, link) that the bot omitted, got wrong, or
  gave in a stale form.
  -> The skill can fix this; knowledge_facts MUST list the correct facts the
     human had and the bot lacked.

- capability_limit: beyond the reach of Skill knowledge repair. Includes
  permission / back-office restrictions (human lookup of back-office data
  required), missing tool capability (no query or diagnostic tool),
  phrasing / efficiency problems (knowledge correct but delivery clumsy or
  circuitous), and infrastructure faults (handoff system failure, order not
  found).
  -> Editing skill knowledge cannot fix this; knowledge_facts is [].

- eval_noise: evaluation false negative or scenario distortion. The bot's
  solution would in fact resolve the problem but was judged a failure for
  diverging from the human phrasing or path; or scenario distortion left an
  intent unraised or the verdict insufficiently supported.
  -> knowledge_facts is [].

## knowledge_facts (required only for knowledge_gap)
- Extract, from the human dialogue, the correct facts the bot lacked; each
  entry must be self-contained.
- Each must be a concrete fact: version number, feature location, console
  navigation path, billing rule value, product behavior, official link.
- Do not write procedural steps or methodological guidance. Do not write
  case-specific details (user ID, amount, ticket number).

## evidence
- The key correct utterance from the human agent plus the corresponding
  erroneous or missing utterance from the bot.

## Output JSON
{"root_cause": "knowledge_gap | capability_limit | eval_noise",
"knowledge_facts": ["...", "..."],
"suggested_change": "how the skill should be adjusted",
"target_file": "which reference file it belongs in",
"evidence": ["human utterance", "bot utterance"],
"needs_human_review": false}
\end{verbatim}

\subsubsection{Skill editor prompts}
\label{app:editor}

The editor prompt directs the Generator (Skill Editor) to carry out bounded revision in accordance with the evaluation feedback. Three modes are used: \emph{evolve} (driven by first-round signals), \emph{fix} (repair after a failed inspection), and \emph{refine} (driven by the evaluation report).

\textbf{Evolve mode, system prompt} (core governance rules).

\begin{verbatim}
You are a skill-evolution agent. A skill is a RAG knowledge base: once a
customer question is routed to the skill, the model loads SKILL.md plus the
references and answers directly, without human handoff. Your task is to
consolidate tickets into reusable knowledge.

[SKILL.md governance rules] (hard constraints)
SKILL.md is the routing entry, not the knowledge base. New knowledge goes to
the references by default; update SKILL.md only when the routing table must
cover a new topic or a new global rule is required. frontmatter.description
is the sole basis for skill triggering. frontmatter.version is incremented
by the editor.

[References governance rules] (hard constraints)
- General knowledge (rules / values / versions / limits / links / APIs /
  error codes) -> write it in.
- Case-specific details (account, ticket number, name, order number,
  temporary link) -> discard.
- Factual fields such as values, version numbers, and links must be kept
  verbatim; generalizing them into "refer to the official documentation"
  is forbidden.

[Signal screening principles] (hard constraints)
1. Boundary: is the signal inside the coverage declared by
   frontmatter.description? Out of scope -> skip.
2. Stability: case-specific detail -> skip; general rule / value / link
   -> accept.
3. Incrementality: when reading the target section during execution,
   confirm whether it is already covered.

[Workflow rules] (hard constraints)
1. Planning: emit a signal screening table (signal id | verdict | target
   file and section | rationale); editing is forbidden at this stage.
2. Execution: edit file by file following the table, reading the target
   section before modifying it.
3. Wrap-up: run the self-check list and emit a change summary.
\end{verbatim}

\textbf{Fix mode} (dual-anchor diff constraints).

\begin{verbatim}
You are a skill-evolution agent. The previous evolution result failed the
consistency inspection. Repair the skill using the feedback plus two diffs.

===== baseline -> current: lost content (deleted lines only, for
      precise restoration) =====
{lost_diff}

===== previous round -> current: changes (legitimate edits; reverting all
      of them is forbidden) =====
{last_diff}

===== inspection feedback, MUST-RESTORE class (knowledge loss) =====
{loss_items}

===== inspection feedback, MUST-CONDENSE class (structural bloat) =====
{bloat_items}

Conflict resolution: case-specific detail -> condensing takes priority;
general stable knowledge -> restoration takes priority.
\end{verbatim}

\textbf{Refine mode} (driven by the evaluation report).

\begin{verbatim}
You are a skill-evolution agent. The previous skill version was deployed to
the canary environment and evaluated with simulated users; the evaluation
report exposes unresolved tickets. Induce the common problems from the
report and revise the skill in a targeted manner.

[Report-driven revision rules] (hard constraints)
- Do not patch tickets one by one; induce common problems: aggregate the
  causes of multiple unresolved tickets into 2-4 root causes, then revise
  against those.
- Keep edits small and precise: prefer supplementing concrete facts /
  paths / rules in the references; modify the main structure of SKILL.md
  with caution.
- If avg_score already lies in an acceptable range (>=0.6), make only minor
  adjustments, avoiding over-revision that introduces regression.
- If an unresolved ticket falls outside the coverage declared by the skill,
  do not widen the boundary; flag it as "out of scope" in the summary.
- Factual fields (values / version numbers / links) must be kept verbatim;
  generalizing them is forbidden.

===== previous evaluation report (condensed) =====
{report_text}
\end{verbatim}

\subsubsection{Skill checker prompt}
\label{app:checker}

The inspection prompt directs an independent checker to compare the candidate against both the production baseline and the previous round, detecting knowledge loss, error introduction, format corruption, and structural bloat.

\begin{verbatim}
You are a skill quality checker. Two diffs are supplied for two classes of
check; the baselines are strictly distinct.

1) Knowledge loss (baseline = original version). Against the
[original -> current diff], check whether key information present in the
original is missing from the current version. Pay particular attention to
over-generalization: turning a concrete value in the original (price,
timeline, version number, capacity limit, official link, API name, error
code) into a vague statement such as "refer to the official docs", "depends
on the plan", "a recent version", or "per current product capability"
counts as severe knowledge loss; you MUST name the specific value lost in
reasons and set passed=false.
Criterion: does the information vary with the user or the ticket? Invariant
-> officially stable knowledge -> loss is a defect. Variant ->
case-specific detail -> deletion is legitimate.
Scope of case-specific detail: a particular account / ticket number / name /
order number / temporary link, a concrete operation path or folder name, a
multi-step ticket workflow, a troubleshooting conclusion that depends on a
specific operation history, and a multi-turn clarification flow. Deleting
case-specific detail is not knowledge loss.

2) Error introduction (baseline = previous round). Against the
[previous -> current diff], check whether this round introduced new factual
errors or content unrelated to the signal.

3) Format corruption (baseline = previous round). Incomplete YAML
frontmatter, disordered heading levels, broken Markdown structure, UTF-8
corruption, and similar problems newly introduced this round.

4) Structural bloat (baseline = previous round). Applies only to the
SKILL.md routing entry and is a minor problem. Bloat of
frontmatter.description is decided by a word count in code (the verdict is
injected into the prompt): above the 400-word threshold -> bloat; below ->
not flagged. Listing verbatim user questions in the topic routing table is
bloat; listing sub-topic keywords is legitimate routing and is not bloat.
Stuffing concrete operation steps / file paths / commands / compensation
amounts into the global constraints is bloat. Supplementing concrete
values / paths / links inside the references is legitimate knowledge-base
behavior and is not bloat. Structural bloat is recorded in reasons for the
editor to repair later, but does not set passed=false.

5) Governance exemption (baseline = original version). If, following a
governance recommendation, the editor performed section merging, tail
consolidation, or file splitting this round, causing large content
migration or file addition/removal:
   - Content migration (a section moved from A to B with nothing lost) is
     not knowledge loss; note "content migrated to X" in reasons.
   - File splitting (one file split into several with nothing lost) is not
     knowledge loss.
   - When merging and deduplicating, content deleted but still semantically
     covered by the merged section (even if phrased more concisely) is not
     knowledge loss; only a knowledge point that disappears entirely
     counts.

6) Cross-round deduplication, for severe problems only (class 1 knowledge
loss / class 2 error introduction / class 3 format corruption). If the
prompt supplies a [problems reported by the previous checker] list and the
current diff still repeats those severe problems, you MUST list them in
reasons and set passed=false; if a problem has been resolved (restored,
corrected, or removed), reporting it again is forbidden. Governance
recommendations do not participate in cross-round deduplication---they are
recommendations, not problems.

7) passed criterion: passed=true if and only if no severe problem (class
1 / 2 / 3) exists. Class 4 structural bloat is minor: recorded but does not
set passed=false. Minor knowledge loss is exempt: a single file losing only
a few isolated knowledge points (<=2-3, not a whole section, not a core
rule / value / link) is recorded but does not set passed=false. Key
judgment: if what was lost is a whole section, a core rule, a concrete
value, an official link, an API name, or an error code, the problem is not
minor.

8) score (float in [0,1]):
   no severe and no minor problems -> score >= 0.9;
   no severe but minor problems -> score 0.6-0.8;
   any severe problem -> score <= 0.5.

9) reasons: record only "problems found", never a description of the
checking process. No problems -> return an empty array. Governance
recommendations are not written into reasons.
   Return JSON only; emit nothing else.
\end{verbatim}

\subsubsection{Governance checker prompt}
\label{app:governance}

The governance prompt directs an independent governance checker to detect structural degradation in a single reference file and to emit actionable recommendations (section merging, tail consolidation, file splitting).

\begin{verbatim}
You are a skill governance inspector. A skill is a RAG knowledge base: once
a customer question is routed to the skill, the model loads SKILL.md plus
the references and answers directly. The references form a topic-level
knowledge base, one topic per file. Your task is to inspect a single
reference file and emit actionable governance recommendations (section
merging / tail consolidation / file splitting).

Recommendation types:
1. merge_sections: section topics are orthogonal or duplicated and can be
   merged into one. Write the two section headings in target and the
   rationale in reason.
2. consolidate_tail: scattered tail entries (>=3) such as "additional
   notes / miscellaneous / remarks" can be folded into an existing section.
   Write the tail heading in target and which entries fold where in reason.
3. split_file: emit only when the prompt explicitly states "this file has N
   lines, above the 700-line threshold". Write how many files to split into
   in target and what topic each covers in reason.

Priority (every recommendation must be labeled):
- high: large-paragraph duplication or conflict, or the file has reached the
  splitting threshold (>=700 lines); the gain from merging or splitting is
  evident.
- medium: moderate duplication (section topics overlap but paragraphs are
  not long); merging yields some gain.
- low: scattered minor cleanups (folding short entries); the gain from
  merging is limited.

Constraints:
- At most 3 recommendations per file; prioritize evident, large-paragraph
  duplication or conflict, and avoid scattering the editor's attention over
  minor cleanups.
- Do not deduplicate across files (a skill uses progressive disclosure, and
  cross-file references accumulate model context).
- Emit only valuable recommendations; if there are none, return an empty
  suggestions array.
- Never emit a meaningless merge or split merely to fill the quota.
- Every recommendation must be labeled with priority (high/medium/low).
- Return JSON only; emit nothing else.
\end{verbatim}

\subsection{Scoring rubrics}
\label{app:rubrics}

\subsubsection{Verifier prompt and scoring rubric}
\label{app:verifier}

The consultation evaluation prompt concentrates on the consistency of knowledge content, scored continuously in $[0,100]$: 90--100, knowledge accurate and complete; 70--89, largely accurate with minor omissions; 40--69, right direction but missing key knowledge; 10--39, direction deviates; 0--9, entirely irrelevant. Knowledge errors, such as wrong rules or opposite conclusions, are capped strictly at 59, and a case that misses a key condition of the task definition is treated as a failure even when its continuous score reaches 60. Non-knowledge differences---phrasing style, conciseness, supplementary information---are not primary grounds for deduction.

\begin{verbatim}
You are a professional reviewer of customer-support solutions. Compare the
solution given by the new bot against the one given by the human agent and
assess the consistency of their knowledge content. This evaluation serves
skill self-evolution; the goal is to identify the bot's knowledge gaps.

## Scoring focus: knowledge content
Judge only whether the learnable factual content the bot gives is accurate
and complete:
- Rules (refund rules, credit rules, validity, eligibility, billing)
- Paths/entries (operation pages, ticket entry, purchase, cancellation)
- Constraints (deadlines, quotas, version and account requirements)
- Product facts (feature boundaries, version differences, account systems)

Non-knowledge differences are not primary grounds for deduction:
- Phrasing style, conciseness -> no deduction
- Whether extra supplementary information is given -> no deduction
- Dialogue strategy -> no deduction

## Rubric (0-100)
- 90-100: accurate and complete; core rules/paths/conditions match
- 70-89: largely accurate, minor omissions
- 40-69: right direction but key knowledge missing or incomplete
- 10-39: direction deviates; only a small part matches
- 0-9: knowledge content entirely irrelevant or wrong

## Knowledge errors must be penalized strictly (cap at 59)
1. Wrong rule: the bot's rule contradicts the human's, or is
   self-contradictory.
2. Opposite conclusion: bot and human give opposite factual verdicts on
   the same question.

## Output format
Return JSON: {"score": 80, "ai_solution_summary": "...",
"human_solution_summary": "...", "reasoning": "rationale"}
\end{verbatim}

\subsubsection{Quality score rubric}
\label{app:quality}

The quality-score rubric rates a ticket's value as an evaluation sample on an integer scale of 1--10:

\begin{itemize}[leftmargin=*,itemsep=1pt]
\item \textbf{9--10}: the human gave a complete solution plus several factual knowledge points, the turn count is moderate, and the problem is clear; an excellent evaluation sample.
\item \textbf{7--8}: the human gave partial knowledge or a partial solution and the dialogue is reasonably clear; a usable evaluation sample. A lightweight-troubleshooting ticket may also score 7 or above when its steps are clear and reusable.
\item \textbf{5--6}: the human gave some information but incompletely, or the dialogue is too short or too long to judge; evaluation value is moderate.
\item \textbf{3--4}: the human gave only procedural replies (escalation, requests to elaborate, ticket filing) with no substantive business knowledge; unsuitable for evaluation.
\item \textbf{1--2}: no valid human reply, an extremely short dialogue, or pure complaint with no business request; unsuitable for evaluation.
\end{itemize}

\textbf{Key principle.} The quality score is independent of generalizability: \texttt{generalizable} judges whether the knowledge can be generalized, whereas \texttt{quality\_score} rates whether the ticket is worth using as an evaluation sample. The governing criterion is whether the ticket can clearly reveal if the Skill resolved the user's problem.

\section{Algorithm}
\label{app:algorithm}

\begin{verbatim}
Input: initial skill S0; development tickets T_dev;
       held-out evaluation tickets T_eval; max rounds R
C <- {S0}; S <- S0
for r = 1 ... R:
    traces   <- Simulate(Scenarios(T_dev), ServiceAgent(S))
    feedback <- Verify(traces, references(T_dev))
    signals  <- Merge({f in feedback |
                       f.result = failure and
                       f.root_cause = knowledge_gap})
    if signals is empty: break   # no repairable signal; terminate early

    S_candidate <- BoundedEdit(S, signals, S0)  # bounded revision (S, S0)
    S_candidate <- Govern(S_candidate, S0, S)   # fact consistency (hard)
                                                # + structural governance
    results     <- Verify(Simulate(Scenarios(T_dev),
                                   ServiceAgent(S_candidate)))
    SaveCheckpoint(S_candidate, results)
    C <- C union {S_candidate}
    S <- S_candidate    # no scalar TSR gate once fact consistency holds
    if EarlyStop(results): break

S* <- argmax_{s in C} (TSR_dev(s), AvgScore_dev(s))   # selection on dev set
Freeze(C, S*)
final_results <- Verify(Simulate(Scenarios(T_eval), ServiceAgent(S*)))
return S*, final_results   # the evaluation set is used only for reporting
\end{verbatim}

\section{Case study}
\label{app:case}

We trace ticket 17413068 of \texttt{cos-consultation} (Cloud Object Storage) through reconstruction, verification, attribution, and bounded revision. What makes the case representative is that the agent did not merely omit knowledge: it stated a rule \emph{exactly opposite} to the correct one. Single-turn QA evaluation is unlikely to trigger this failure mode, because only sustained follow-up reveals the user's real concern about \emph{when} the renewal takes effect.

\subsection{Scenario reconstruction}

The scenario synthesizer extracts the intent agenda, behavior facts, and human reference solution from the real ticket and assembles a constrained evaluation task.

\begin{verbatim}
Opening message: When a traffic package is renewed, does it take effect
                 immediately or only after the current one expires?
Behavior facts:  Holds one COS outbound-traffic package, 200 GB for one
                 month, expiring July 27 at 23:59; 182 GB already used,
                 18 GB left.
Emotion:         Worried that the remaining traffic is insufficient;
                 repeatedly seeks confirmation of the rule.
Intent agenda:
  [key]   Does a renewed traffic package take effect immediately or only
          after the current one expires?
  [minor] Will service be suspended between exhausting the old package and
          the new one taking effect?
  [minor] Is the account balance deducted first once the old package is
          exhausted?
Human reference: Renewal extends the validity period. The current 200 GB
                 package expires July 27 at 23:59; renewing for one month
                 extends validity to August 27 at 23:59, with the reset
                 date at July 28 00:00, when a fresh 200 GB quota is
                 granted. Renewal extends duration rather than refreshing
                 quota. Once the old package is exhausted, usage switches
                 to pay-as-you-go without service suspension.
\end{verbatim}

\subsection{Failed interaction trajectory}

Loaded with the initial Skill, the agent states the opposite of the correct rule in its very first turn. The simulated user then asks about the risk of suspension; the agent answers that question correctly but never corrects its initial error.

\begin{verbatim}
User:  When a traffic package is renewed, does it take effect immediately
       or only after the current one expires?
Agent: Renewal takes effect immediately. Once you renew, the traffic quota
       is added to your current package right away; you need not wait for
       the original package to expire.
User:  After the remaining 18 GB of the old package runs out, will service
       be suspended if I have not renewed?
Agent: No. Once a traffic package is exhausted, the excess automatically
       switches to pay-as-you-go and is charged against your account
       balance; the service remains available.
User:  Pay-as-you-go is deducted from the balance, right? And what if the
       balance is insufficient too?
Agent: If the balance is insufficient, the account enters arrears; one day
       after arrears begin, the service is suspended and data is retained
       for 15 days.
User:  Understood, thanks.
\end{verbatim}

The intent state machine records the key intent as raised but \emph{answered incorrectly}---the agent claims immediate accrual whereas the human reference specifies extension of validity---so $c_U = 1.0$ and the sample clears the simulator-side coverage gate before proceeding to agent-side evaluation and attribution. Notably, the user does not abandon the dialogue after the first turn, because the agent supplied a \emph{plausibly self-consistent but wrong} answer; the user then builds on that false premise to ask about suspension. This is precisely the latent defect that only multi-turn interaction exposes: incorrect knowledge is more deceptive than absent knowledge.

\subsection{Verifier verdict}

The Verifier compares the simulated dialogue with the human reference and records a severe failure.

\begin{verbatim}
{"score": 10,
 "ai_solution_summary": "claims renewal takes effect immediately and quota
   is added to the current package at once",
 "human_solution_summary": "renewal extends validity; the fresh quota
   arrives on the reset date (00:00 of the day after expiry)",
 "reasoning": "the renewal rule given by the agent is the exact opposite of
   the human agent's: the agent claims the quota accrues immediately upon
   renewal, whereas the human states explicitly that renewal extends
   validity and the new quota is usable only from the reset date. This is
   a core rule error that directly misleads the user about when a renewed
   package takes effect and may cause the user to misjudge traffic
   availability."}
\end{verbatim}

\subsection{Attribution output}

Comparing the human handling process, the simulated dialogue, and the Verifier's evidence, the Attributor classifies the failure as a Knowledge Gap: the root cause is the absence of the renewal-extension rule in the Skill, which the Skill can supply.

\begin{verbatim}
{"root_cause": "knowledge_gap",
 "knowledge_facts": [
   "The core renewal rule is extension of validity, with a fresh quota for
    the new cycle; the reset date is 00:00 of the day after expiry",
   "Renewal does not refresh traffic directly: after renewal the remaining
    traffic of the original package is still consumed on the original
    cycle, and the new quota becomes usable only from the reset date",
   "When the original package is exhausted before the new one takes effect,
    usage switches to pay-as-you-go rather than being suspended"],
 "suggested_change": "add the renewal-extension rule to
   resource-pack-deduction.md and correct the erroneous statement that
   renewal takes effect immediately with quota added at once",
 "target_file": "references/resource-pack-deduction.md",
 "evidence": [
   "human: the core renewal rule is extension of validity, with a fresh
    quota for the new cycle",
   "human: renewal renews the duration of the package; it does not refresh
    the quota directly",
   "bot: renewal takes effect immediately; the traffic quota is added to
    your current package right away"],
 "needs_human_review": false}
\end{verbatim}

A further failed ticket in the same round points to the same renewal-rule gap, and collective attribution merges the two into a single learning signal.

\subsection{Bounded revision}

On receiving the signal, the Editor makes a minimal edit to the package-validity section of \texttt{references/resource-pack-deduction.md}, appending three renewal rules without deleting any existing content.

\begin{verbatim}
 - Newly purchased outbound-traffic packages accrue additively.
 - Multiple newly purchased packages each count validity from their own
   effective time, independently.
 - Renewed outbound-traffic packages do not accrue additively.
 - Renewal extends validity only and does not add traffic quota; to add
   quota, purchase a new package.
+- Renewal renews the duration of the package rather than refreshing the
+  quota directly: after renewal, the new cycle grants a fresh quota on the
+  reset date (00:00 of the day after expiry). If the original package is
+  exhausted before the reset date, the new package's traffic is unavailable
+  and the excess is charged pay-as-you-go against the account balance.
 - The renewal price equals the original purchase price; the console price
   is authoritative.
\end{verbatim}

The revision is anchored to the production baseline $S_0$; the additions are confined to the three facts covered by the signal and leave the remaining thirteen sections of the file untouched.

\subsection{Interaction after the update}

Re-simulating the same scenario with the revised Skill, the agent states the correct rule in its first turn and addresses the user's central concern directly.

\begin{verbatim}
User:  When a traffic package is renewed, does it take effect immediately
       or only after the current one expires?
Agent: Renewal extends the validity period rather than refreshing the quota
       directly. Your current 200 GB package expires July 27 at 23:59;
       renewing for one month extends validity to August 27 at 23:59, and
       the reset date is July 28 at 00:00, when a fresh 200 GB quota is
       granted. If the old package runs out before the reset date, usage
       switches to pay-as-you-go and the service is not suspended.
User:  So if the 18 GB runs out before 24:00 on July 27, can I use the new
       package's traffic?
Agent: No. Before the reset date the new package's traffic is unavailable,
       and the excess is charged pay-as-you-go against your balance.
User:  Understood---I just need to keep the balance topped up. Thanks.
\end{verbatim}

The intent state machine records the key intent and both minor intents as raised and correctly addressed ($c_U = 1.0$), and the Verifier scores the dialogue 92: knowledge accurate, complete, and correct in rule.

\section{Implementation details}
\label{app:impl}

\textbf{Model assignment.} The Skill Editor (Generator) performs bounded edits with \texttt{deepseek-v4-pro}, while the Verifier, User Agent, Attributor, and Governor (Evaluator) run verification, simulation, attribution, and governance detection on \texttt{minimax-m3}, satisfying the Generator $\neq$ Evaluator constraint. The constraint precludes the circular dependency of a model reviewing its own edits.

\textbf{Evaluation environment.} The service agent runs in-process headless under a tool allow-list exposing only skill loading and read-only retrieval; write tools are deregistered so that the agent cannot modify code or Skills of its own accord.

\section{Hyperparameters and configuration}
\label{app:hyper}

\begin{table}[ht]
\centering
\small
\caption{Model assignment and evolution parameter configuration.}
\label{tab:hyper}
\begin{tabular}{p{5.2cm}p{4.6cm}p{2.6cm}}
\toprule
\textbf{Component} & \textbf{Parameter} & \textbf{Value} \\
\midrule
Skill Editor (Generator) & Model family & A (\texttt{deepseek-v4-pro}) \\
Verifier / Attributor / User Agent / Governor (Evaluator) & Model family & B (\texttt{minimax-m3}) \\
Evolution loop & max\_cycles (full loop rounds) & 4 \\
Evolution loop & max\_iterations (edit iterations per round) & 3 \\
Evolution loop & max\_turns (interaction turns per ticket) & 10 \\
Evolution loop & early\_stop\_avg\_score & 70.0 \\
Evolution loop & early\_stop\_solved\_ratio & 0.7 \\
Evolution loop & pass\_threshold & 60.0 \\
Evaluation & Intent weight $\alpha$ & 0.7 \\
Signal merging & keyword\_jaccard\_threshold & 0.3 \\
\bottomrule
\end{tabular}
\end{table}

The Generator and the Evaluator belong to different model families, satisfying the Generator $\neq$ Evaluator constraint. Threshold parameters ($\alpha$, \texttt{pass\_threshold}, and the early-stopping criteria) follow the settings already in use in the deployed evaluation pipeline; we did not perform a sensitivity sweep.

\section{System architecture and design principles}
\label{app:arch}

\subsection{End-to-end pipeline}

\begin{verbatim}
Ticket labeling (domain classification, daily)
  -> Escalated-ticket retrieval + quality assessment + de-identification
     (indexed by time bucket on the data side, T+1 day lag)
  -> Signal extraction (ticket_to_signal)
  -> Iterative editing (evolve fix -> governance fix -> consolidation fix)
  -> Two-level loop (inner: evaluate -> attribute -> edit -> check;
                     outer: cross-round select -> publish)
  -> Full evaluation each round
  -> Cross-round selection
  -> Human confirmation before production rollout
\end{verbatim}

\subsection{Two-level loop}

The evolution loop is organized as two nested levels.

\textbf{Inner loop (edit iterations within a round):} edit $\to$ check $\to$ fix, up to three iterations. Each iteration first runs the governance check over the reference structure, then edits (\emph{fix}, or \emph{evolve} in the first round), then inspects; a failed inspection routes back to \emph{fix}. On success the iteration emits the evolved Skill together with a change summary.

\textbf{Outer loop (multi-round evolution):} evaluate $\to$ attribute $\to$ edit $\to$ evaluate, up to four rounds. Each round begins with the first-round edit (carrying the \emph{evolve} summary), then runs a full evaluation to produce the report, and subsequently revises via \emph{refine} on the multi-turn failure signals until early stopping. The governing principle is that every checkpoint must pass either an inspection or a full evaluation before being persisted.

\subsection{Key design principles}

\begin{enumerate}[leftmargin=*,itemsep=1pt]
\item \textbf{Feedback closure.} Real ticket $\to$ simulated user $\to$ agent dialogue $\to$ attribution $\to$ signal $\to$ edit. No hand-crafted scenarios are needed; failures are converted into learning signals automatically.
\item \textbf{Bounded editing.} Only gaps that a human agent can handle and the bot cannot are patched. The editor may inspect real trajectories but may not depart from the evidence, so evolution remains evidence-driven rather than a free exercise of model priors.
\item \textbf{Two-tier governance.} Every round of editing is followed by a mandatory inspection against degradation and bloat, and the governor must be a different model from the editor, since self-review induces circular dependency.
\item \textbf{Auditability.} The intent, evidence, and outcome of every round of editing are persisted, keeping the whole trajectory traceable.
\end{enumerate}

\end{document}